\documentclass[conference]{IEEEtran}
\IEEEoverridecommandlockouts
\usepackage{geometry}

\newgeometry{left=54pt, right=54pt, top=72pt, bottom=54pt}

\usepackage{cite}
\usepackage{amsmath,amssymb,amsfonts}
\usepackage{algorithmic}
\usepackage{graphicx}
\usepackage{caption}
\usepackage{textcomp}
\usepackage{xcolor}
\usepackage[export]{adjustbox}
\usepackage{url}
\usepackage{balance}
\usepackage{academicons}
\definecolor{orcidlogocol}{HTML}{A6CE39}
\definecolor{lime}{HTML}{A6CE39}
\DeclareRobustCommand{\orcidicon}{%
    \begin{tikzpicture}
    \draw[lime, fill=lime] (0,0) 
    circle [radius=0.16] 
    node[white] {{\fontfamily{qag}\selectfont \tiny ID}};
    \draw[white, fill=white] (-0.0625,0.095) 
    circle [radius=0.007];
    \end{tikzpicture}
    \hspace{-2mm}
}

\newcommand{\orcidWalter}{\href{https://orcid.org/0000-0003-4565-1272}{\orcidicon}}
\newcommand{\orcidRoss}{\href{https://orcid.org/0000-0001-8595-0379}{\orcidicon}}
\newcommand{\orcidXingcheng}{\href{https://orcid.org/0000-0003-1178-5221}{\orcidicon}}
\newcommand{\orcidRui}{\href{https://orcid.org/0000-0001-7359-1081}{\orcidicon}}
\newcommand{\orcidMarc}{\href{https://orcid.org/0009-0008-9223-2015}{\orcidicon}}
\newcommand{\orcidDaniel}{\href{https://orcid.org/0009-0008-5346-5473}{\orcidicon}}
\newcommand{\orcidAhmed}{\href{https://orcid.org/0000-0003-3702-8042}{\orcidicon}}
\newcommand{\orcidAkshay}{\href{https://orcid.org/0009-0000-4969-0119}{\orcidicon}}
\newcommand{\orcidTrivedi}{\href{https://orcid.org/0000-0002-0937-6771}{\orcidicon}}
\newcommand{\orcidKnoll
}{\href{https://orcid.org/0000-0003-4840-076X}{\orcidicon}}

\usepackage{tikz}

\usepackage{hyperref} 
\hypersetup{
    bookmarks=false,
    colorlinks=true,
    breaklinks=true,
    linkcolor=blue,
    filecolor=magenta,      
    hypertexnames=true,
    urlcolor=cyan,
    pagebackref=true,
    pdftitle={ICRA40},
    pdfpagemode=FullScreen,
    citecolor=green
    }
\def\BibTeX{{\rm B\kern-.05em{\sc i\kern-.025em b}\kern-.08em
    T\kern-.1667em\lower.7ex\hbox{E}\kern-.125emX}}
\begin{document}
\restoregeometry

\title{Safety-Critical Learning for Long-Tail Events: \\ The TUM Traffic Accident Dataset}
%\title{The TUM Traffic Accident Dataset}
%\title{Overcoming Challenges of Long-Tail Events in Safety Critical Domains: a Framework for Deriving Efficient, Robust Models with a Novel Real-World Traffic Public Dataset}

\author{
Walter Zimmer$^1$~\orcidWalter \quad
%\IEEEauthorblockN{Walter Zimmer}
%\IEEEauthorblockA{\textit{TU Munich}
%}
%\and
Ross Greer$^{2,5}$~\orcidRoss \quad
%\IEEEauthorblockN{Ross Greer}
%\IEEEauthorblockA{
%\textit{UC San Diego}}
%\and
Xingcheng Zhou$^1$~\orcidXingcheng \quad
%\IEEEauthorblockN{Xingcheng Zhou}
%\IEEEauthorblockA{
%\textit{TU Munich}}
%\and
Rui Song$^{1,3}$~\orcidRui \quad
%\IEEEauthorblockN{Rui Song}
%\IEEEauthorblockA{
%\textit{TU Munich}}
%\and
Marc Pavel$^1$~\orcidMarc \quad\\
%\IEEEauthorblockN{Marc Pavel}
%\and
Daniel Lehmberg$^1$~\orcidDaniel~
%\IEEEauthorblockN{Daniel Lehmberg}
%\and
Ahmed Ghita$^{1,4}$~\orcidAhmed
Akshay Gopalkrishnan$^{2}$~\orcidAkshay~
%\IEEEauthorblockN{Ahmed Ghita}
%\IEEEauthorblockA{
%\textit{TU Munich}}
%\and
Mohan Trivedi$^2$~\orcidTrivedi~
%\IEEEauthorblockN{Mohan Trivedi}
%\IEEEauthorblockA{
%\textit{UC San Diego}}
%\and
Alois Knoll$^1$~\orcidKnoll

%\IEEEauthorblockN{Alois Knoll}
%\IEEEauthorblockA{TU Munich}
\thanks{This research was supported by the Federal Ministry of Education and Research in Germany within the $\text{\textit{AUTOtech.agil}}$ project (GN: 01IS22088U).}
\thanks{$^1$ W. Zimmer, X. Zhou, R. Song, M. Pavel, D. Lehmberg, A. Ghita, and A. Knoll are with the Chair of Artificial Intelligence and Robotics at the Technical University of Munich.}
\thanks{$^2$ R. Greer, A. Gopalkrishnan and M. Trivedi are with the Laboratory for Intelligent and Safe Autombiles (LISA) at the Uni. of California San Diego.}
\thanks{$^3$ R. Song is with the Chair of Artificial Intelligence and Robotics at the Technical University of Munich and with Fraunhofer Institute for Transportation and Infrastructure Systems (IVI).}
\thanks{$^4$ A. Ghita is with SETLabs Research GmbH.}
\thanks{$^5$ R. Greer is with the University of California Merced.}
}

%\addtolength{\topmargin}{50pt}

%\maketitle
\makeatletter
\let\@oldmaketitle\@maketitle
\renewcommand{\@maketitle}{\@oldmaketitle
\setcounter{figure}{0}    
  %\vspace{-0.9cm}
  % left, bottom, right, top
  \centering
  \url{https://tum-traffic-dataset.github.io/tumtraf-a}\\[5pt]
  \includegraphics[width=0.49\linewidth,frame,trim={0cm 4cm 0cm 1cm},clip]{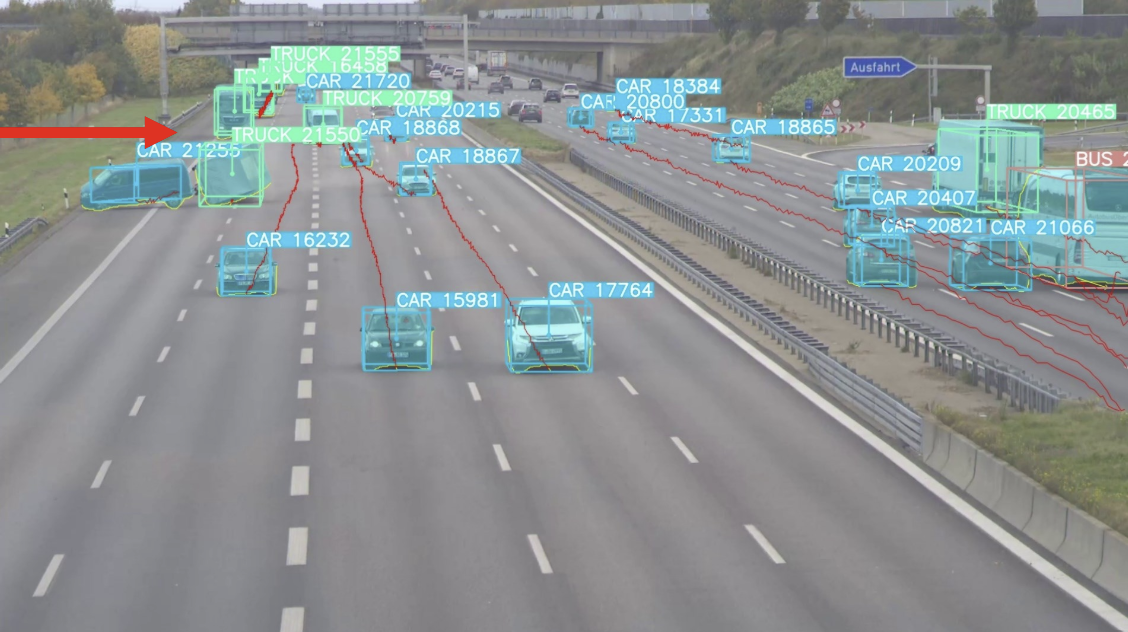}
  \includegraphics[width=.49\textwidth,frame,trim={0cm 15.2cm 5.5cm 0cm},clip]{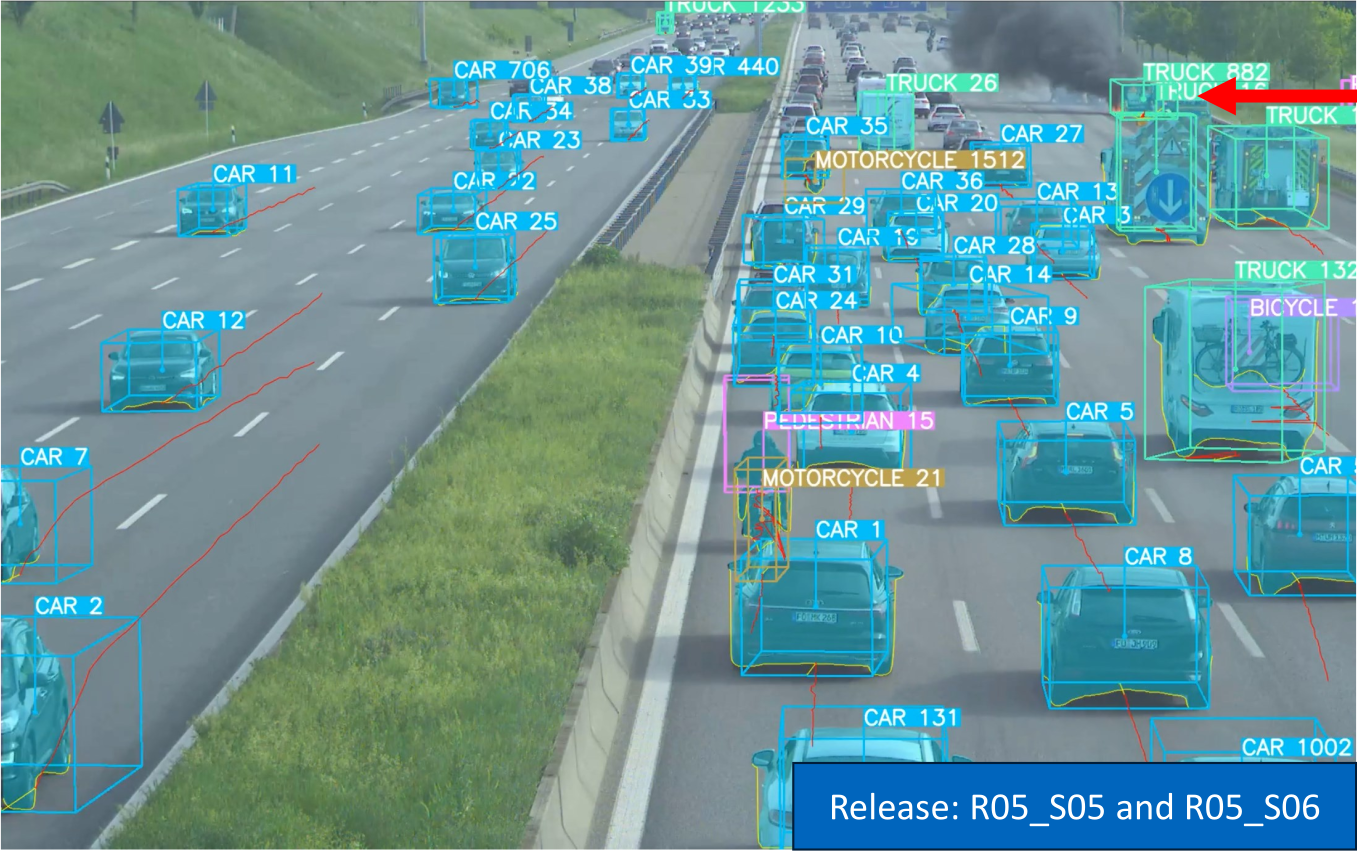}
  %\bigskip
  %\vspace{-0.2cm}
  \captionof{figure}{\textbf{Visualization of the \textit{TUM Traffic Accident} dataset} with 3D box annotations, track IDs and trajectories. Accidents are recorded from four roadside cameras on the \textit{A9 Test Bed for Autonomous Driving} in Munich, Germany. Left: a vehicle is in the process of overturning following collision. Right: a vehicle has pulled over after catching fire following a collision.}
  \label{fig:overview_figure}
  }
\makeatother

\maketitle

% 8655 + 289 = 8944
\begin{abstract}
Even though a significant amount of work has been done to increase the safety of transportation networks, accidents still occur regularly. They must be understood as an unavoidable and sporadic outcome of traffic networks. We present the \textit{TUM Traffic Accident} (TUMTraf-A) dataset, a collection of real-world highway accidents. It contains ten sequences of vehicle crashes at high-speed driving with 294,924 labeled 2D and 93,012 labeled 3D boxes and track IDs within 48,144 labeled frames recorded from four roadside cameras and LiDARs at 10 Hz. The dataset contains ten object classes and is provided in the \textit{OpenLABEL} format. We propose \textit{Accid3nD}, an accident detection model that combines a rule-based approach with a learning-based one. Experiments and ablation studies on our dataset show the robustness of our proposed method. The dataset, model, and code are available on our project website.
\end{abstract}

% \begin{IEEEkeywords}
% component, formatting, style, styling, insert
% \end{IEEEkeywords}

\section{Introduction}
%: Long-Tail Events in Autonomous Driving}

Collecting data on events that occur most rarely (``long-tail events") is important for robust machine learning of data-driven perception, planning, and control models in robotic systems \cite{christianos2023planning}. In the case of autonomous driving, however, these events are especially costly to collect \cite{ghita2024activeanno3d, kulkarni2021create, fingscheidt2022deep}. Long-tail events such as accidents and near-misses \cite{kataoka2018drive} come at great risk to human life and are otherwise difficult to stage and capture in the natural driving environment \cite{wang2022ips300+, greer2024and}. 
The time between the occurrence of an accident and the arrival of medical assistance significantly impacts whether the passengers of a vehicle survive an accident. Automatic accident detection systems reduce this time and have the potential to save lives. 

Our dataset supports the development of safer autonomous systems through multiple directions. First, the data supports improved learning on a variety of perception-related tasks, such as detection \cite{liu2024graphrelate3d,carta2024roadsense3d,zimmer2023infradet3d,zimmer2023real,zimmer2022survey,zimmer2022realdomain,philipsen2015traffic,abualsaud2021laneaf}, tracking, and segmentation \cite{greer2024patterns}. Second, the nature of the dataset allows for the study of methods of cooperative perception between roadside sensors observing the same scene from different view angles. Cooperative perception allows for a reduction of occlusion through shared information. Third, the introduction of roadside sensors allows for the creation of digital twins of the traffic scene \cite{kremer2023digital}. These digital twins expand the visibility range beyond the egocentric view, which may be used to provide adequate warning lead time for safe approaches and control transitions for risky traffic events \cite{greer2023safe}.

\begin{figure*}[t]
    \centering
    \includegraphics[width=0.49\linewidth,trim={0cm 1cm 0cm 3cm},clip,frame]{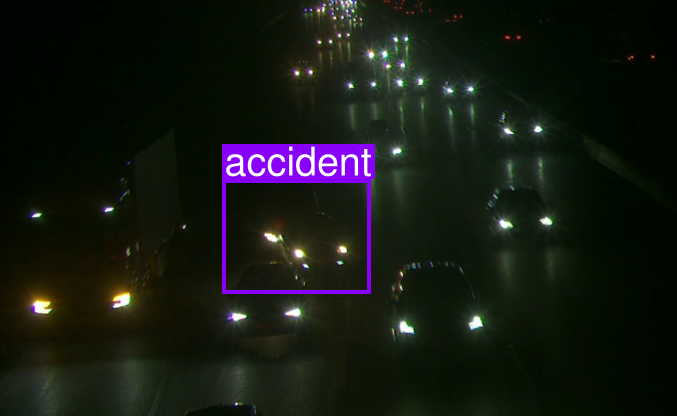}
    \includegraphics[width=0.49\linewidth,trim={0cm 1cm 0cm 3.3cm},clip,frame]{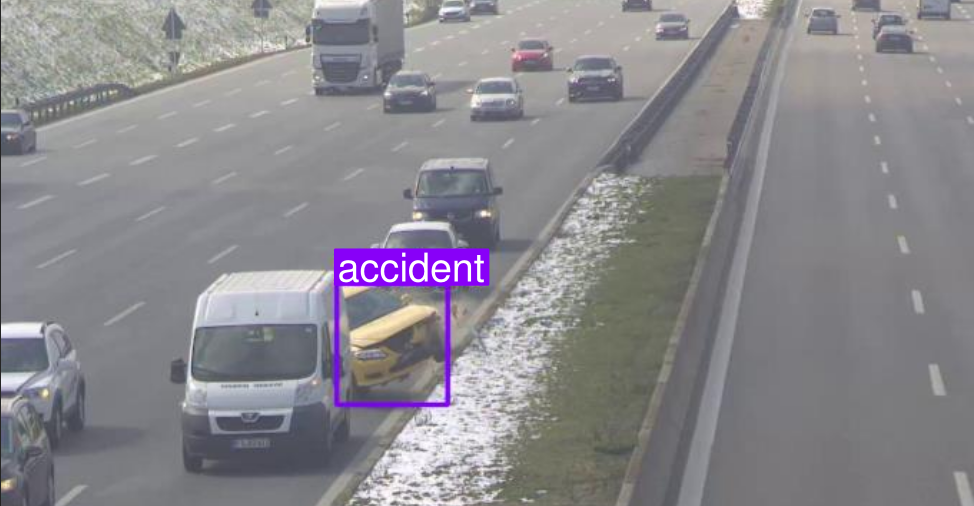}
    \caption{\textbf{Qualitative visualization results of our accident detection framework on the \textit{TUM Traffic Accident} test set.}\\ Left: The rule-based approach detected a rear-end collision. Right: The learning-based approach detected a car crash.}
    \label{fig:qualitative_results}
\end{figure*}

% [] TODO: extend contribution
\textbf{Our main contributions are:}
\begin{itemize}
\item We present the \textit{TUM Traffic Accident} dataset, a dataset curated specifically for these rare and dangerous occurrences, with 294,924 2D and 93,012 3D annotations.  
\item We introduce \textit{Accid3nD}, a framework that can detect and analyze accidents and near-miss events in real-time and in different weather and lighting conditions.
\item Experiments and ablation studies show that our method achieves state-of-the-art results on our dataset.
\item We open source our dataset, accident detection framework and development kit on our project website.
\end{itemize}

\section{Related Work}
Existing accident detection methods have never been tested on real traffic data of a test stretch. Real accident datasets are rare and do not contain enough data to train deep learning models \cite{liu2024survey}.
The \textit{DeepAccident} \cite{wang2024deepaccident} dataset contains 691 synthetic accident scenarios in the CARLA simulator. These accidents were generated based on crash reports published by the \textit{National Highway
Traffic Safety Administration} (NHTSA). The dataset contains labeled data from four vehicles and one roadside infrastructure camera. One limitation of this work is that all accidents are generated in a simulation environment and do not represent realistic crash scenes. Hence, the sim-to-real gap needs to be addressed to improve the generalization capabilities of the perception models.

\section{The TUM Traffic Accident Dataset}
% TODO: [] check whether we need to mention all other datasets (-> consumes 1/4 page)
The \textit{TUM Traffic Accident} dataset is part of the larger family of TUM Traffic Datasets, including the \textit{TUM Traffic A9 Highway} dataset \cite{cress2022tumtrafa9}, the \textit{TUM Traffic Intersection} dataset \cite{zimmer2023tumtraf}, the \textit{TUM Traffic Event} dataset \cite{cress2024tumtrafevent}, and the \textit{TUM Traffic V2X Cooperative Perception} dataset \cite{zimmer2024tumtraf}.

% [] TODO: describe TUM Traffic datasets in full paper version.
% \begin{itemize}
%     \item \textit{TUM Traffic A9 Highway} dataset \cite{cress2022tumtrafa9}, which contains labeled multi-sensor data with a mix of random and sequential traffic scenarios from the A9 autobahn, including extreme weather such as rain, wind, snow, and fog,  
%     \item \textit{TUM Traffic Intersection} dataset \cite{zimmer2023tumtraf}, which contains traffic at a busy intersection performing complex driving maneuvers such as left and right turns, overtaking, and U-turns, 
%     \item \textit{TUM Traffic Event} dataset \cite{cress2024tumtrafevent}, which contains synchronized image material between an Event-Based and an RGB camera, and 
%     \item \textit{TUM Traffic V2X Cooperative Perception} dataset \cite{zimmer2024tumtraf}, which contains labeled point clouds, images, and GPS and IMU data from both roadside and onboard sensors for cooperative 3D object detection and tracking in occlusion scenarios, including events like traffic violations and near misses. 
% \end{itemize}

It features 48,144 labeled camera and LiDAR frames with 294,924 2D and 93,012 3D box annotations, tracking labels and trajectory information, and classification of ten different instance types, including cars, trucks, buses, trailers, vans, pedestrians, motorcycles, bicycles, emergency vehicles, and others. The data is captured from roadside cameras and LiDARs during day and nighttime and labeled with the \textit{3D Bounding Box Annotation Toolbox} (3D BAT) \cite{zimmer20193d}. Featured accidents include instances of high-speed lane changes with failure to notice stopped traffic, overturning of vehicles during collision, vehicles catching fire, a variety of emergency response vehicles, and more. Some example cases are illustrated in Fig. \ref{fig:overview_figure}.

The data from the \textit{TUM Traffic Accident} dataset can be used as ground truth for the development and verification of AI-based detectors, tracking, fusion algorithms, trajectory prediction, and to understand and analyze the occurrence and the after-effects of naturally occurring high-speed crash incidents and other accidents on the autobahn.

The \textit{TUM Traffic Accident} dataset, as well as its preceding datasets \cite{zimmer2024tumtrafdatasets}, are available for academia and industry and include a development kit repository for ease of use \cite{zimmer2024devkit}.

%\section{Creating Safer Autonomous Systems}

\section{Accident Detection}
We propose a method to automatically detect accidents on the highway in real-time using roadside cameras. 
Our method consists of a rule-based and a learning-based approach. We first use a rule-based approach that uses vehicle trajectories as input to detect accidents based on predefined thresholds. This method outputs an accident classification for every vehicle in the current frame in real-time.

If an accident is detected, the learning-based approach is executed to make a final prediction of whether or not an accident is present in the camera image. We train a \textit{YOLOv8} model on our accident dataset to detect accidents in an image. These detections are then filtered by a score threshold of 0.8, and an accident needs to be detected in at least three consecutive frames to reduce the number of false positive detections. Furthermore, we fuse the accident detection results of all cameras available in a recording of a driving scenario.

We recorded camera images and the fused perception results for 128 days, stored them into rosbag files, and processed these recordings. 
The automatic accident analysis was executed on 12,290 15-minute videos.
In total, 831,969 unique vehicles were identified. We detected 3,748 standing vehicles in a driving lane, 138 standing vehicles in a shoulder lane, and 120 breakdown events. Qualitative results of the rule-based and learning-based accident detection are shown in Fig. \ref{fig:qualitative_results}. Related approaches may be used to detect both accidents and general traffic anomalies \cite{greer2024towards, greer2023pedestrian}. 

\section{Evaluation}
We evaluate the accuracy and runtime performance of both accident detection models on our \textit{TUM Traffic Accident} dataset. 
The rule-based approach runs with 10.41 ms per frame (95.05 FPS) on an NVIDIA RTX 3090 GPU. The total processing runtime for a 15-minute long rosbag with 22,500 ROS messages recorded at 25 FPS is 234.25 seconds.

\section{Conclusion}
In this work, we present the \textit{TUM Traffic Accident} dataset and evaluate two accident detection methods on real data. One limitation of our rule-based approach is that it can only detect rear-end collisions, which will be improved in future work.

%\section{Acknowledgment}
%This research was supported by the Federal Ministry of Education and Research in Germany within the $\text{\textit{AUTOtech.agil}}$ project, Grant Number: 01IS22088U.

% \clearpage
% \pagebreak

%\addtolength{\textheight}{-20cm}
\balance
\bibliographystyle{ieeetr}
\bibliography{main}

\end{document}